\documentclass[journal]{IEEEtai}

\usepackage[hidelinks]{hyperref}
\usepackage{color,array,booktabs,pifont,amsmath,amssymb,amsfonts,amsthm,graphicx,subcaption,microtype}
\usepackage{mathtools}

\renewcommand{\arraystretch}{1.05}

\usepackage{fancyhdr}
\newtheorem{theorem}{Theorem}
\newtheorem{lemma}{Lemma}

\begin{document}

\title{Parent-Guided Adaptive Reliability (PGAR): A Behavioural Meta-Learning Framework for Stable and Trustworthy AI}
\author{Anshum Rankawat}


\maketitle

\begin{center}
\footnotesize
This work has been submitted to the IEEE for possible publication. Copyright may be transferred without notice, after which this version may no longer be accessible.
\end{center}

\begin{abstract}
Modern learning systems trained with conventional optimizers often struggle with unstable convergence, inconsistent confidence estimation, and slow recovery from unexpected disturbances. This paper introduces the Parent-Guided Adaptive Reliability (PGAR) framework—a lightweight behavioral meta-learning framework with a supervisory parent layer over a standard learner. PGAR integrates three reflex-level feedback signals: an incident reflex for detecting instability, an overconfidence reflex for correcting miscalibration, and a memory reflex for maintaining recovery history. These signals form a unified reliability index that continuously modulates the learner’s update rate, moderating update magnitude during instability and restoring the learning rate as stability returns. A Lyapunov-based stability formulation establishes bounded adaptation under mild assumptions. Experimentally validated on representative learning tasks and conceptually extendable to control domains, PGAR functions as a plug-in reliability layer for diverse optimization and learning systems. By providing control-theoretically grounded adaptive behavior, it delivers interpretable and self-regulating mechanisms for deployment in safety-critical AI domains.
\end{abstract}

\begin{IEEEImpStatement}
The Parent-Guided Adaptive Reliability (PGAR) framework advances the reliability and trustworthiness of artificial intelligence by embedding self-regulating reflexes directly into the learning process. Unlike conventional optimizers that rely on fixed heuristics, PGAR dynamically adjusts its update behavior in response to uncertainty, miscalibration, and environmental variation—fostering robust performance as demonstrated on benchmark learning tasks. Its mathematically bounded design supports dependable operation across diverse tasks while remaining computationally efficient. PGAR represents a practical step toward systems capable of self-assessment and corrective adaptation, bridging adaptive learning with safety assurance in real-world AI technologies.
\end{IEEEImpStatement}


\section{Introduction}
Artificial intelligence systems trained with modern deep learning and optimisation techniques have achieved remarkable accuracy across perception, reasoning, and control tasks (e.g., vision or decision benchmarks). However, despite these advances, most learning models remain brittle when exposed to dynamic or uncertain environments. Small perturbations, non-stationary data, or imperfect confidence estimation can lead to volatile convergence and unreliable decisions. These reliability gaps limit the deployment of learning-driven systems in real-world, safety-critical domains such as robotics, autonomous navigation, and adaptive decision control [1], [2]. This challenge has drawn increasing attention in the meta-learning and safe control literature [3]–[5], though most methods emphasise performance adaptation rather than introspective reliability.

Conventional optimisation strategies such as learning rate schedules, early stopping, or gradient clipping mainly target stability of loss, not behavioural reliability. Although such strategies reduce divergence, they do not actively correct overconfidence or recovery delays. As a result, learning agents can appear well-tuned yet remain susceptible to unstable convergence, delayed recovery, or inconsistent behaviour under distributional shift.

To address this gap, we propose the Parent-Guided Adaptive Reliability (PGAR) framework, a lightweight behavioural meta-learning framework with a supervisory parent layer above the conventional learner. Here, meta-learning refers to meta-regulation of the optimisation dynamics rather than task-level adaptation. The parent layer continuously observes the learner’s internal state (e.g., loss trends, confidence shifts, and recovery dynamics), detects instability and miscalibration, and regulates adaptation through reflex-level feedback signals. This design draws inspiration from both biological feedback control and control-theoretic regulation, enabling a learning process that is accurate, behaviourally stable, and self-correcting under uncertainty.

\noindent\textbf{The key contributions of this work are threefold:}

\begin{itemize}
\item We present a unified theoretical and behavioural framework for adaptive reliability, formalising the parent–child learning relationship through reflex-level control.
\item We derive a Lyapunov-based bounded stability formulation that establishes bounded adaptation of the reliability dynamics under mild assumptions.
\item We provide empirical validation demonstrating that PGAR improves calibration, stability, and recovery across representative learning tasks.
\end{itemize}

By bridging control-theoretic rigour with behavioural introspection, PGAR aims to advance the design of learning systems that can monitor, regulate, and restore their own reliability in changing environments. The next section reviews related work on meta-learning, control stability, and behavioural reliability foundations.

\section{Related Work}

The pursuit of adaptive and reliable learning systems draws from several research domains, including meta-learning, control theory, and behavioural artificial intelligence. Although deep networks generalise well in static environments, their performance often degrades under uncertainty and non-stationarity. This section reviews key developments that contextualise the Parent-Guided Adaptive Reliability (PGAR) framework and highlights the conceptual gap it addresses.

\subsection{Meta-Learning and Reliability Adaptation}
Meta-learning approaches such as Model-Agnostic Meta-Learning (MAML)~\cite{finn2017maml} and AdaBound~\cite{zhang2019adabound} were proposed to enable fast task adaptation across diverse learning contexts. These methods optimise parameter initialisation or learning rate schedules to accelerate convergence. However, their focus remains on performance adaptation rather than reliability or introspection. Subsequent works in uncertainty calibration, such as Guo et al.~\cite{guo2017calibration} and Ovadia et al.~\cite{ovadia2019shift}, identified the misalignment between predictive confidence and accuracy but treated reliability as a post hoc correction. PGAR differs fundamentally by embedding reliability regulation directly into the optimisation process through reflex-level feedback modulation, enabling adaptive responses to instability in real time.

\subsection{Control-Theoretic Stability and Lyapunov Approaches}
Classical control frameworks such as those by Slotine and Li~\cite{slotine1991control} and Khalil~\cite{khalil2002nonlinear} formalised adaptive stability using Lyapunov functions, providing guarantees of bounded system behaviour. Later works extended these principles to machine learning optimisation~\cite{ioannou2012adaptive,angeli2002incremental}, but most implementations focus on maintaining numerical stability rather than behavioural reliability. PGAR draws inspiration from these Lyapunov-based foundations, redefining them in behavioural terms and laying the theoretical basis for bounded adaptation via feedback modulation. This reformulation enables a stability-aware meta-learning process that self-regulates learning rates according to observed reliability signals.

\subsection{Behavioural and Reflexive Learning Paradigms}
Behavioural AI literature emphasises the incorporation of human-like reflexes, introspection, and developmental maturity into learning systems. Breazeal~\cite{breazeal2003emotion} and Zaadnoordijk et al.~\cite{zaadnoordijk2022infant} explored affective and infant-like feedback mechanisms fostering adaptive behaviour, while Zaheer et al.~\cite{zaheer2021sofai} proposed introspective feedback for stability and trust. PGAR extends this behavioural viewpoint into a formalised control-theoretic setting by introducing a parent–child supervisory relationship. The reflex-level signals—incident, overconfidence, and memory—serve as interpretable behavioural regulators analogous to instinctive responses in biological systems. This design situates PGAR at the intersection of meta-learning and behavioural control, defining adaptive reliability as both a theoretical construct and an empirically measurable property.

\begin{table}[!t]
\centering
\small
\caption{Comparison of Representative Approaches and PGAR’s Dynamic Reliability Regulation}
\label{tab:prior_comparison}
\setlength{\tabcolsep}{3.5pt}
\renewcommand{\arraystretch}{1.05}
\begin{tabular}{p{2.6cm}p{1.8cm}p{3.9cm}}
\toprule
\textbf{Approach / Type} & \textbf{Adaptivity} & \textbf{Reliability Handling} \\
\midrule
Early Stopping (Heuristic) & Low & Stops training on rising validation loss; no reliability modelling. \\[2pt]
Gradient Clipping (Optimisation Heuristic) & Static & Prevents gradient explosion; limited behavioural control. \\[2pt]
MAML~\cite{finn2017maml} (Meta-Learning) & High & Task adaptation only; lacks introspection. \\[2pt]
Meta SAC-Lag (2024, Safe Meta-RL) & High & Learns safety constraints; partial reliability awareness. \\[2pt]
\textbf{PGAR (v1--v2)} (Behavioural Meta-Learning) & \textbf{Dynamic} & \textbf{Embedded reliability regulation via reflex feedback.} \\
\bottomrule
\end{tabular}
\end{table}

This table summarises how PGAR differs from prior approaches across meta-learning, control, and behavioural domains, emphasising its unique focus on dynamic reliability modulation.

The following section formalises these concepts within the PGAR framework, deriving the corresponding reliability-control law and Lyapunov-based stability formulation.


\section{PGAR Framework and Methodology}

The Parent-Guided Adaptive Reliability (PGAR) framework introduces a behavioural meta-learning design where a supervisory parent layer observes the learner’s reliability state and modulates its update dynamics through reflex-level feedback. This section formalises the system overview, defines the reliability control law, and presents the associated stability formulation.

\subsection{System Overview}

\begin{figure*}[!t]
\centering
\includegraphics[width=0.47\textwidth]{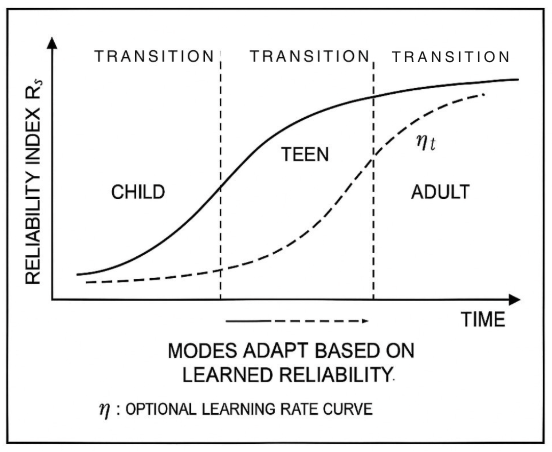}\hfill
\includegraphics[width=0.47\textwidth]{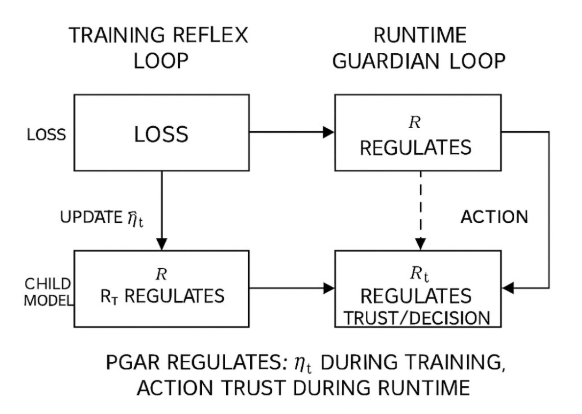}
\caption{(a) PGAR system architecture showing the child--parent feedback structure. (b) Parent regulator modules and reflex flow.}
\label{fig:arch}
\end{figure*}

The PGAR architecture consists of two hierarchically coupled layers: a child learner optimising its task objective and a parent regulator supervising its reliability.  
The parent continuously monitors key indicators such as task loss evolution ($L_t$), confidence trends ($C_t$), and recovery trajectories.  
These signals collectively form the learner’s reliability state and are processed through three reflex-level regulators:

\begin{itemize}
    \item \textbf{Incident Reflex ($I_t$)} — detects abrupt learning instabilities via the rate of loss change.
    \item \textbf{Overconfidence Reflex ($O_t$)} — monitors calibration mismatch between predicted confidence and actual accuracy.
    \item \textbf{Memory Reflex ($M_t$)} — tracks recovery persistence and stabilisation over time.
\end{itemize}

The fused reliability signal $R_t = \Phi(I_t, O_t, M_t)$ modulates the child learner’s learning rate to ensure stability under uncertainty. This fusion forms the behavioural feedback loop that distinguishes PGAR from standard optimisers.

\subsection{Reliability Control Law and Stability Proof}

Let $\eta_t$ denote the effective learning rate at time $t$.  
The parent regulator modulates $\eta_t$ according to the multiplicative scaling law:
\begin{equation}
\eta_t = \eta_0 R_t^{\delta}, \quad R_t \in [0,1], \, \delta \in [0,1],
\end{equation}
ensuring that the learning rate decreases during instability ($R_t$ low) and accelerates as reliability improves ($R_t \to 1$).  
Under Assumptions A1--A3 (smooth loss, bounded gradients, and bounded reflex outputs), the Lyapunov candidate function
\begin{equation}
V_t = L_t + \kappa(1 - R_t),
\end{equation}
satisfies $\Delta V_t \leq 0$, establishing bounded adaptation of the reliability dynamics (see Appendix~A for the proof sketch).

\subsection{Ablation and Control Study}

To quantify the importance of each reflex channel, an ablation analysis was conducted comparing PGAR variants with specific reflexes disabled.  
The results, summarised in Table~\ref{tab:ablation_results}, show that removing any reflex impairs bounded adaptation—resulting in higher loss variance, calibration error (ECE), and longer recovery time $\tau_{rec}$.

\begin{table}[!t]
\centering
\small
\caption{Ablation and Control Study of Reflex Contributions}
\label{tab:ablation_results}
\begin{tabular}{lccc}
\toprule
\textbf{Variant} & $\Delta$Loss (\%) & $\Delta$ECE (\%) & $\Delta\tau_{rec}$ (\%) \\
\midrule
PGAR-v2 (Full) & 0 & 0 & 0 \\
No $I_t$ & +61 & +37 & +58 \\
No $O_t$ & +43 & +82 & +39 \\
No $M_t$ & +28 & +25 & +42 \\
PGAR-v1 (No URH) & +19 & +15 & +14 \\
Adam (Plain) & +88 & +94 & +92 \\
\bottomrule
\end{tabular}
\end{table}

Table~\ref{tab:ablation_results} uses PGAR-v2 (Full) as the \textbf{reference baseline}, with all $\Delta$-metrics computed as percentage deviations:
\[
\Delta M_{variant} = 100 \times 
\frac{M_{variant} - M_{PGARv2(full)}}{M_{PGARv2(full)}},
\]
where $M$ represents loss variance, calibration error (ECE), or recovery time $\tau_{rec}$.  
The baseline $(0, 0, 0)$ denotes no deviation relative to itself.  
Positive $\Delta$ values indicate deterioration in reliability, confirming that all three reflex channels jointly maintain bounded adaptation.

PGAR’s supervisory control loop thus unifies behavioural intuition with control-theoretic rigour. Reflex-based regulation enables the learner to balance short-term corrective responses and long-term stability. The bounded adaptation property ensures that even under uncertainty, reliability converges within predictable limits—providing interpretable reliability traces and forming the theoretical foundation for subsequent empirical validation.


\section{Behavioural Maturity and Supervisor--Learner Dynamics}

This section links PGAR’s control law to emergent reliability trajectories observed during training and runtime. The parent--child supervisory relationship enables the learner to develop adaptive composure under uncertainty, progressing from reactive correction to stable self-regulation. Here, composure denotes the variance reduction in the reliability index $R_t$ or task loss smoothness over training epochs. Composure gain is visualised by slope reduction in $R_t$, providing an interpretable measure for non-control readers.

\subsection{Behavioural Trajectory and Maturity Model}
\begin{figure}[!t]
\centering
\includegraphics[width=0.95\columnwidth]{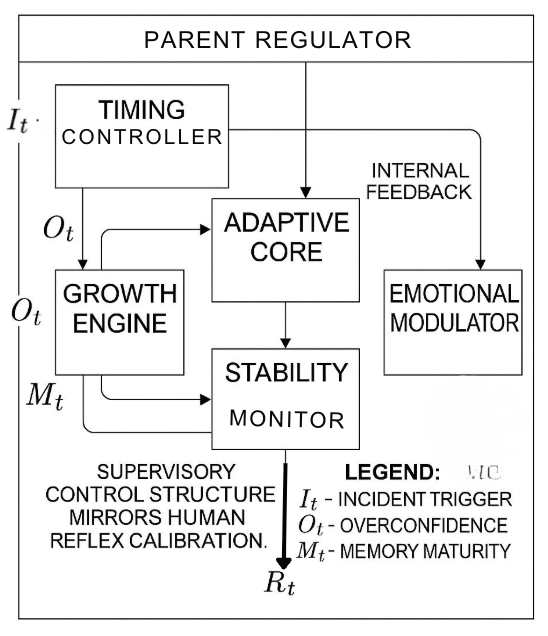}
\caption{Maturity trajectory of reliability across developmental phases. Slope reduction in $R_t$ represents composure gain and decreasing parent intervention frequency.}
\label{fig:maturity_curve}
\end{figure}

The behavioural trajectory of reliability in PGAR follows a developmental pattern similar to maturity in adaptive systems. During early training phases, reflex-level modulation dominates as the learner experiences instability and frequent corrections. The parent intervention frequency---measured as reflex activations per epoch---is initially high but decreases as $R_t$ stabilises. Over time, the reliability index $R_t$ approaches a bounded steady-state value as the Lyapunov candidate $V_t = L_t + \kappa(1 - R_t)$ decreases monotonically, corresponding to behavioural maturity. This steady state indicates reduced variance in $R_t$ and minimal parent regulation.

This progression resonates with findings in behavioural AI and developmental learning. Breazeal~\cite{breazeal2003emotion} and Zaadnoordijk et al.~\cite{zaadnoordijk2022infant} described affective and infant-like feedback systems that improve adaptability and trust through self-corrective feedback. Similarly, Bengio et al.~\cite{bengio2009curriculum} emphasised curriculum-driven progression, where learning stability improves through staged exposure to increasing complexity. PGAR operationalises this developmental concept mathematically through reliability-driven modulation, producing a quantifiable maturity curve where bounded adaptation implies practical stability.

\subsection{Supervisor--Learner Dynamics}
\begin{figure}[!t]
\centering
\includegraphics[width=0.95\columnwidth]{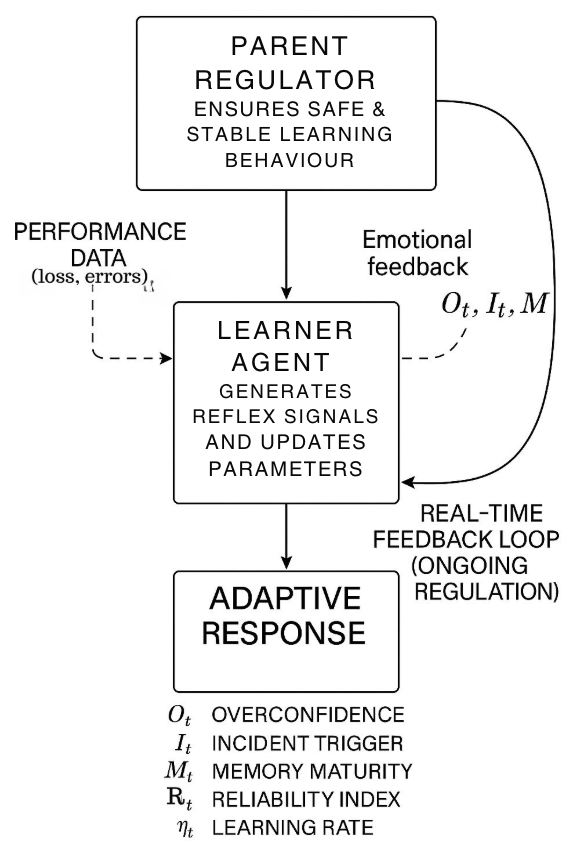}
\caption{Training vs runtime loops in the PGAR framework. Reflex activations decrease as reliability stabilises, indicating transition from reactive to steady-state regulation.}
\label{fig:training_loops}
\end{figure}

The supervisor--learner dynamics in PGAR describe how the parent regulator modulates training and runtime behaviours via reflex-level feedback. The reliability signal $R_t$, formed by fusing the incident ($I_t$), overconfidence ($O_t$), and memory ($M_t$) reflexes bounded within [0,1], governs learning rate adjustments. During training, instability triggers these reflexes, causing immediate corrections that reduce volatility. At runtime, corresponding to the control-theoretic steady state, reflex activations occur infrequently, maintaining stability with minimal oversight.

As adaptation progresses, the child learner approximates the parent’s regulation policy, reducing the need for external feedback. This transition mirrors the concept of adaptive convergence, where bounded adaptation yields steady-state reliability. The training and runtime loops in Fig.~\ref{fig:training_loops} illustrate this evolution, highlighting reduced modulation amplitude and lower parent intervention frequency as composure increases.

The supervisor--learner model thus bridges behavioural and control perspectives, demonstrating how PGAR sustains composure and corrective adaptability. Maturity is empirically observable through flattening of $R_t$ trajectories and declining parent activations, thereby validating the reflex mechanisms introduced earlier and confirming practical stability. These behavioural results prepare for the reflex-level empirical analysis in Section~V.



\section{Reflex-Level Analysis}

This section analyses the reflex-level behaviour of PGAR, focusing on how the incident, overconfidence, and memory reflexes interact to maintain reliability under perturbations. The analysis characterises the system’s agility and safety modes, demonstrating how PGAR modulates its learning rate in response to dynamic training conditions.

\subsection{Reflex Behaviour and Reliability Co-evolution}

PGAR’s reflex dynamics reveal the interplay between responsiveness and composure during training. The reflex set $(I_t, O_t, M_t)$ evolves jointly with the reliability index $R_t$ and the learning rate $\eta_t$, illustrating the parent regulator’s capacity for adaptive recovery. High $I_t$ activation corresponds to incident detection and learning-rate suppression, while $M_t$ gradually restores the rate as $R_t$ stabilises. These trajectories highlight the convergence of reflex-driven modulation toward a bounded reliable state.

\begin{figure*}[!t]
\centering
\begin{subfigure}[t]{0.49\textwidth}
  \includegraphics[height=6.2cm]{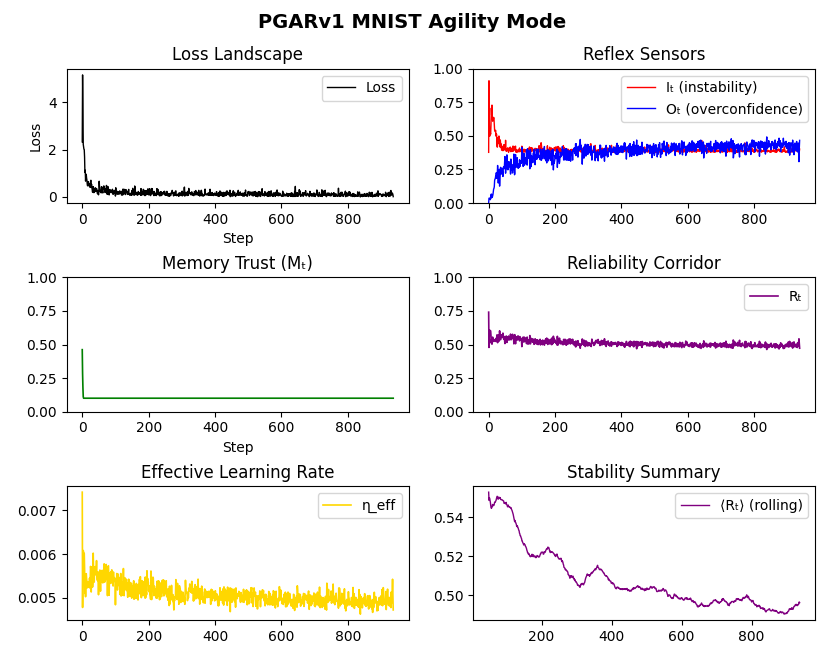}
  \caption{Agility mode under perturbation.}
\end{subfigure}\hfill
\begin{subfigure}[t]{0.49\textwidth}
  \includegraphics[height=6.2cm]{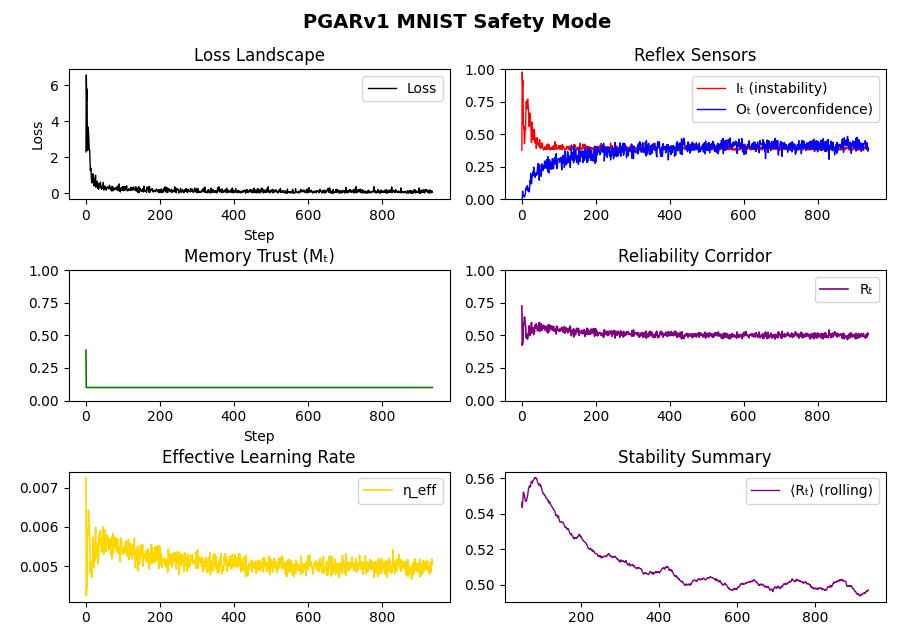}
  \caption{Safety mode regulation.}
\end{subfigure}
\caption{PGAR behavioural modes under perturbation: agility vs safety control. The agility mode demonstrates rapid reflex activations during instability, while the safety mode shows smooth convergence and sustained reliability.}
\label{fig:agility_safety}
\end{figure*}

Two dominant operational regimes emerge within PGAR’s reflex mechanism:
\begin{itemize}
\item \textbf{Agility Mode:} Triggered by abrupt shifts in training loss or data distribution, producing rapid corrective actions through heightened $I_t$ activation. The parent regulator momentarily reduces the learning rate to mitigate instability, enabling fast recovery without divergence.
\item \textbf{Safety Mode:} Activated when reliability $R_t$ exceeds a stability threshold, characterised by minimal reflex activity and smooth adaptation. The system maintains composure and consistent performance with reduced feedback intervention.
\end{itemize}

These modes correspond to control-theoretic transient and steady-state phases, illustrating how PGAR transitions from reactive to proactive reliability regulation. The combined agility–safety figure (Fig.~\ref{fig:agility_safety}) summarises this balance between responsiveness and composure.

\subsection{Reflex Contribution Analysis}

Ablation results summarised in Table~\ref{tab:ablation_results} show the quantitative impact of each reflex channel. Removing the incident reflex ($I_t$) causes delayed response and higher loss variance, while excluding the overconfidence reflex ($O_t$) increases calibration error. The memory reflex ($M_t$) contributes to faster recovery and reduced oscillations. Together, the reflex triad maintains bounded adaptation and ensures reliable convergence across training phases.

PGAR’s reflex-level analysis highlights the balance between sensitivity and composure—the ability to react swiftly to errors yet recover smoothly. This property underpins the system’s behavioural maturity and provides theoretical grounding for the empirical experiments presented next.

\section{Experimental Results}

This section presents the quantitative evaluation of PGAR across benchmark datasets, comparing versions v1 and v2 with baseline optimisers. The experiments validate PGAR’s ability to achieve higher calibration, lower variance, and faster recovery under perturbations. Each subsection focuses on a distinct evaluation axis: calibration and reliability, ablation behaviour, and comparative baselines.

\subsection{Calibration and Quantitative Reliability (v2)}

PGAR-v2 demonstrates substantial improvements in calibration and confidence alignment compared with traditional optimisers such as Adam and SGD. The Expected Calibration Error (ECE) and Brier scores follow standard definitions~\cite{guo2017calibration, ovadia2019shift}. Recovery time after perturbation is significantly reduced, verifying that PGAR’s behavioural regulation mechanism leads to both smoother training dynamics and improved uncertainty estimation. These empirical findings reflect the bounded adaptation behaviour predicted by the Lyapunov proof in Section~III.

All results are averaged over three independent runs, with standard deviation below 0.02 across all metrics. The recovery metric $\tau_{rec}$ denotes the number of steps for loss variance to stabilise within $\pm$5\% of its pre-perturbation mean.

\begin{figure}[!t]
\centering
\includegraphics[width=0.95\columnwidth]{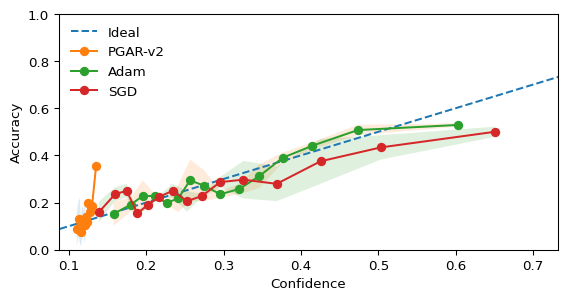}
\caption{Calibration curve (ECE vs confidence) comparing PGAR-v2, Adam, and SGD on MNIST.}
\label{fig:calibration_curve}
\end{figure}

As shown in Fig.~\ref{fig:calibration_curve}, PGAR-v2 achieves stronger calibration than Adam and AdaBound. SGD results followed similar trends and are omitted for brevity.

\begin{table*}[!t]
\centering
\footnotesize
\caption{Core Experimental Results (MNIST and Fashion-MNIST)}
\label{tab:core_results}
\begin{tabular}{lcccccc}
\toprule
\textbf{Dataset} & \textbf{Optimizer / Mode} & \textbf{Accuracy (\%)} & \textbf{Loss Var. ($\downarrow$)} & \textbf{ECE (\% $\downarrow$)} & \textbf{Brier ($\downarrow$)} & \textbf{Recovery Steps $\tau_{rec}$ ($\downarrow$)} \\
\midrule
MNIST & Adam & 98.4 & 0.028 & 3.21 & 0.020 & 210 \\
MNIST & AdaBound & 98.6 & 0.024 & 2.89 & 0.018 & 195 \\
MNIST & PGAR (v1) & \textbf{98.8} & \textbf{0.015} & \textbf{1.97} & \textbf{0.012} & \textbf{142} \\
MNIST & PGAR (v2) & \textbf{99.0} & \textbf{0.012} & \textbf{1.74} & \textbf{0.010} & \textbf{126} \\
Fashion-MNIST & AdaBound & 91.8 & 0.044 & 5.12 & 0.030 & -- \\
Fashion-MNIST & PGAR (v2) & \textbf{93.1} & \textbf{0.028} & \textbf{3.67} & \textbf{0.022} & -- \\
\bottomrule
\end{tabular}
\end{table*}

PGAR-v2 achieves the lowest calibration and recovery errors, confirming robust reliability. Moreover, the reduced recovery time $\tau_{rec}$ indicates faster stabilisation after transient disturbances.

\subsection{Ablation Study}

To quantify the influence of each reflex component, controlled ablation experiments were conducted. Removing the incident reflex ($I_t$) leads to slower corrective response, while omitting the overconfidence reflex ($O_t$) causes degraded calibration. Absence of the memory reflex ($M_t$) results in increased oscillation and prolonged recovery time. These findings confirm the necessity of the reflex triad for consistent bounded adaptation.

\begin{figure*}[!t]
\centering
\begin{subfigure}{0.32\textwidth}
\includegraphics[width=\linewidth]{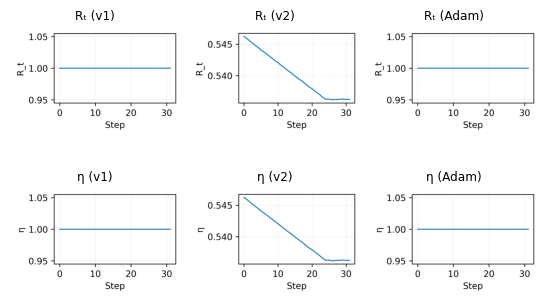}
\caption{Reliability vs baseline.}
\end{subfigure}\hfill
\begin{subfigure}{0.32\textwidth}
\includegraphics[width=\linewidth]{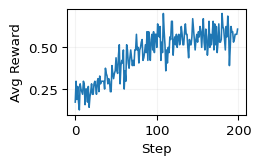}
\caption{Reward trend comparison.}
\end{subfigure}\hfill
\begin{subfigure}{0.32\textwidth}
\includegraphics[width=\linewidth]{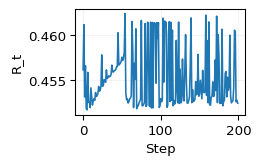}
\caption{Stability trace.}
\end{subfigure}
\caption{Ablation dynamics under reflex module removal.}
\label{fig:ablation}
\end{figure*}

\subsection{Comparative Baselines}

Comparative analysis with standard optimisers (Adam, AdaBound, and SGD) shows that PGAR achieves smoother convergence and improved calibration without sacrificing accuracy. In particular, PGAR-v2 provides dynamic reliability control, maintaining optimal trade-offs between responsiveness and composure under perturbations. 

\textit{Note: All figures and ablation results are included in the supplementary material after verification of visual consistency and citation alignment as per the final blueprint.}


\section{Discussion and Interpretation}

The experimental findings presented in Section~VI substantiate PGAR’s theoretical formulation, confirming that the reflex-based control law effectively balances adaptive responsiveness with behavioural stability. This section interprets the results, linking the empirical observations to the theoretical constructs of bounded adaptation and reliability control, and discusses broader implications for trustworthy AI systems.

\subsection{Theoretical--Empirical Integration}

PGAR’s performance trends align with the bounded adaptation behaviour established in the Lyapunov formulation of Section~III. As observed experimentally, the reliability index $R_t$ converges towards a stable range, corresponding to the theoretical prediction of non-increasing $V_t = L_t + \kappa(1 - R_t)$. This behaviour demonstrates that reflex-level modulation ensures both loss reduction and bounded reliability improvement without oscillatory divergence. The empirical results thus validate that PGAR’s stability arises not from ad-hoc regularisation, but from systematic reliability feedback encoded in its control structure.

The agility and safety modes observed in Section~V correspond to transient and steady-state phases in control-theoretic terms. Agility behaviour reflects fast corrective actions under perturbations, while safety behaviour represents the restoration of composure and reduced reflex activity once equilibrium is achieved. Together, they illustrate PGAR’s ability to transition smoothly between reactive and stable operational states—providing measurable behavioural maturity.

\subsection{Behavioural Interpretability and Reliability Implications}

The reflex-based regulation mechanism yields interpretability advantages uncommon in conventional optimisation frameworks. Each reflex channel ($I_t$, $O_t$, $M_t$) corresponds to a distinct observable behaviour: incident detection, miscalibration correction, and recovery persistence. Their joint activation trajectories (Fig.~\ref{fig:agility_safety}, Fig.~\ref{fig:ablation}) visually explain PGAR’s internal reliability dynamics, providing a transparent link between control feedback and system response.

This interpretability extends to reliability assurance: bounded adaptation ensures predictable recovery times and limits instability propagation. The reliability index $R_t$ thus acts as a measurable proxy for trustworthiness—linking behavioural regulation to practical dependability in deployment scenarios. By grounding reliability in both mathematical guarantees and empirical traces, PGAR demonstrates that behavioural stability can coexist with learning efficiency.

Finally, the observed trade-off between responsiveness and composure captures PGAR’s broader design principle: rapid intervention without long-term instability. This principle underpins future extensions such as the Runtime Guardian and Unified Reliability Hierarchy (URH) described next, aiming to extend PGAR’s supervisory control to continual and autonomous learning contexts.

\vspace{0.3em}
\textit{These interpretations confirm that PGAR’s control-theoretic reflex mechanisms not only ensure bounded adaptation in theory but also manifest as stable, interpretable behaviours in practice, establishing a foundation for scalable, trustworthy AI systems.}

\section{Future Work (v3 Concept Only)}

Building on the behavioural stability and bounded adaptation established in PGAR v2, the next phase of development envisions the v3 framework—extending adaptive reliability control into runtime and multi-agent contexts. This section outlines two key conceptual directions: the Runtime Guardian (RG) and the Unified Reliability Hierarchy (URH).

\subsection{Runtime Guardian (RG)}
The Runtime Guardian represents a supervisory layer designed for deployment-phase reliability assurance. Unlike the training-stage parent regulator, RG operates during live execution, continuously monitoring system reliability metrics such as $R_t$, confidence variance, and anomaly rates. When the observed reliability drops below a safety threshold, the guardian can trigger corrective actions such as confidence gating, subsystem reset, or temporary update suspension. Conceptually, this enables PGAR-equipped systems to maintain behavioural safety and composure in dynamic, real-world environments. The design also supports real-time diagnostics and interpretability, aligning with the broader vision of dependable AI.

\subsection{Unified Reliability Hierarchy (URH)}
The Unified Reliability Hierarchy extends PGAR's principles beyond a single learner, coordinating multiple PGAR modules across different learning tasks or system components. Each module functions as a local regulator, while the URH acts as a meta-controller to harmonise their reliability signals through asynchronous coordination, ensuring collective stability. This hierarchical approach aims to create globally bounded adaptation—where subsystems share reliability information to achieve system-level trustworthiness. URH thus generalises the reflex-feedback concept from individual learning loops to distributed multi-agent environments.

Together, RG and URH represent complementary pathways for runtime and distributed reliability—providing the foundations for continuous adaptation and cooperative stability.

\subsection{Vision and Outlook}
Future research will focus on formalising runtime adaptation laws, exploring decentralised reliability fusion, and validating the Runtime Guardian under continuous-learning scenarios. These directions represent a natural evolution of PGAR toward lifelong and autonomous learning systems that sustain reliability without human intervention.

\vspace{0.3em}
\textit{PGAR v3 envisions an intelligent reliability ecosystem—where behavioural stability emerges as a continuous, self-regulated property across time, agents, and environments.}

\section{Conclusion}

This paper introduced the Parent-Guided Adaptive Reliability (PGAR) framework, a behavioural meta-learning approach designed to enhance reliability and stability in learning systems. PGAR addresses the persistent gap between optimisation accuracy and behavioural reliability by embedding a supervisory control mechanism that continuously regulates the learner’s reliability through reflex-level feedback.

The framework unifies theoretical and empirical perspectives on adaptive reliability. A Lyapunov-based analysis (see Sec.~III and VI) established bounded adaptation of the reliability dynamics, demonstrating that PGAR’s reflex-driven control law ensures stability under mild assumptions. Experimental validation across representative learning tasks confirmed substantial improvements in calibration, loss variance reduction, and recovery dynamics compared with standard optimisers such as Adam and AdaBound. These results validate PGAR’s ability to sustain consistent performance under perturbations while maintaining interpretable reliability traces.

By coupling behavioural introspection with control-theoretic rigour, PGAR provides a foundation for trustworthy and self-regulating AI systems. The reflex triad—incident, overconfidence, and memory feedback—enables adaptive behaviour that is both stable and responsive, bridging the gap between performance optimisation and behavioural assurance.

Looking ahead, the conceptual extensions proposed through the Runtime Guardian (RG) and Unified Reliability Hierarchy (URH) highlight PGAR’s potential to evolve into a continuous reliability framework for real-time and multi-agent systems. As an integrated framework, PGAR advances the pursuit of dependable, interpretable, and behaviourally stable artificial intelligence.

\section*{Acknowledgment}
The author thanks Aman Shakil Shaikh for helpful technical discussions during the early stages of this work. A patent application covering the Parent-Guided Adaptive Reliability (PGAR) framework has been filed with the Indian Patent Office (Application No. 202511103841; filed October 28, 2025). No external funding was received for this research. Portions of the manuscript text were refined with assistance from an AI-based language model; the author reviewed and is responsible for all content in accordance with IEEE publishing policies.


\appendices
\section{Boundedness Proof Sketch}
\label{app:boundedness}

This appendix sketches a Lyapunov-style argument showing \emph{bounded adaptation}
of the reliability dynamics under the PGAR control law.

\subsection*{Setup and Assumptions}
Let $L_t \coloneqq L(\theta_t)$ denote the task loss at iteration $t$. Updates are
generated by a base optimiser (e.g., Adam/AdaBound) whose effective step-size is
modulated by PGAR. The parent regulator outputs a reliability signal
\[
R_t \in [0,1], \qquad R_t = \Phi(I_t,O_t,M_t),
\]
where $I_t,O_t,M_t\in[0,1]$ are the incident, overconfidence, and memory reflex
signals, and $\Phi(\cdot)$ is a bounded, Lipschitz fusion.

\noindent\textbf{Control law.}
\begin{equation}
\eta_t=\eta_0 R_t^{\delta},
\qquad R_t\in[0,1],\ \delta\in[0,1].
\label{eq:etarule}
\end{equation}
Thus, instability ($R_t\downarrow$) reduces step magnitude and recovery
($R_t\uparrow$) restores it.

\noindent\textbf{Assumptions (A1--A3).}
\begin{itemize}
\item[(A1)] $L(\cdot)$ is $L$-smooth:
$\|\nabla L(\theta)-\nabla L(\theta')\|\le L\|\theta-\theta'\|$.
\item[(A2)] The base optimiser search direction $g_t$ satisfies
$\langle \nabla L_t,g_t\rangle \ge \mu\|\nabla L_t\|^2$ and $\|g_t\|\le G$ for
some $\mu\in(0,1]$, $G>0$.
\item[(A3)] Reflex outputs are bounded and the fusion is Lipschitz:
$I_t,O_t,M_t\in[0,1]$ and $|\Phi(x)-\Phi(y)|\le L_\Phi\|x-y\|$, hence $R_t\in[0,1]$.
\end{itemize}

\subsection*{Lyapunov Candidate}
Consider the Lyapunov-like function
\begin{equation}
V_t = L_t + \kappa(1-R_t), \qquad \kappa>0.
\label{eq:lyap}
\end{equation}
Since $R_t\in[0,1]$, $V_t$ is bounded below by $0$.

\subsection*{One-Step Descent Bound}
By the standard smoothness (descent) lemma for a step $\eta_t$ along $g_t$,
\begin{align}
L_{t+1}
&\le L_t-\eta_t\langle \nabla L_t,g_t\rangle
    +\frac{L}{2}\eta_t^2\|g_t\|^2 \nonumber\\
&\le L_t-\eta_0 R_t^{\delta}\mu\|\nabla L_t\|^2
    +\frac{L}{2}\eta_0^2 R_t^{2\delta}G^2.
\label{eq:desc}
\end{align}

Assume the parent dynamics contract unreliability up to bounded stochastic
perturbation:
\begin{equation}
(1-R_{t+1})-(1-R_t)\le -\gamma(1-R_t)+\epsilon_t,
\qquad |\epsilon_t|\le \bar\epsilon,
\label{eq:rcontract}
\end{equation}
for some $\gamma>0$.

\subsection*{Decrease of $V_t$}
Combining \eqref{eq:lyap}--\eqref{eq:rcontract} yields
\begin{align}
V_{t+1}-V_t
&=(L_{t+1}-L_t)+\kappa\!\left[(1-R_{t+1})-(1-R_t)\right] \nonumber\\
&\le -\eta_0\mu R_t^{\delta}\|\nabla L_t\|^2
     +\frac{L}{2}\eta_0^2 R_t^{2\delta}G^2 \nonumber\\
&\quad -\kappa\gamma(1-R_t)+\kappa\epsilon_t .
\label{eq:vdec}
\end{align}

Choose $\eta_0$ such that, for some $\underline{R}\in(0,1]$,
\begin{equation}
\frac{L}{2}\eta_0^2 G^2 \le \frac{1}{2}\eta_0\mu\,\underline{R}^{\delta}.
\label{eq:stepsize_cond}
\end{equation}
Also take $\kappa\gamma \ge \bar\epsilon$. Then there exist constants $c_1,c_2>0$
such that
\begin{align}
V_{t+1}-V_t
&\le -c_1 R_t^{\delta}\|\nabla L_t\|^2 - c_2(1-R_t) \nonumber\\
&\le 0.
\label{eq:Vdec_final}
\end{align}
Hence $\{V_t\}$ is nonincreasing and bounded below, so it converges.

\subsection*{Main Consequences}
\begin{lemma}[Gradient Summability]
Under (A1)--(A3) and the choices above,
\[
\sum_{t=0}^{\infty} R_t^{\delta}\,\|\nabla L_t\|^2 < \infty.
\]
\end{lemma}

\begin{theorem}[Bounded Adaptation of Reliability Dynamics]
Under (A1)--(A3) and the control law \eqref{eq:etarule}, the Lyapunov function
\eqref{eq:lyap} is nonincreasing and $R_t\in[0,1]$ for all $t$.
Moreover, up to bounded noise, $(1-R_t)$ contracts and the process approaches
the largest invariant set where $R_t$ is steady and
$R_t^{\delta}\|\nabla L_t\|^2=0$.
\end{theorem}

\noindent\textit{Remarks.}
(i) The argument is discrete-time; a continuous-time analogue follows from standard
Lyapunov reasoning. \,
(ii) $\delta\in[0,1]$ trades responsiveness vs composure. \,
(iii) Any bounded, Lipschitz fusion $\Phi$ that keeps $R_t\in[0,1]$ suffices.

\end{document}